\documentclass[11pt]{article}
\usepackage[margin=1in]{geometry}
\usepackage[T1]{fontenc}
\usepackage{times}
\usepackage{amsmath,amssymb,amsfonts,mathtools}
\usepackage{booktabs,multirow,makecell}
\usepackage{graphicx,subcaption}
\usepackage{xcolor}
\usepackage{tikz}
\usetikzlibrary{arrows.meta,positioning,fit,backgrounds,shapes.geometric,calc}
\usepackage{algorithm}
\usepackage[noend]{algpseudocode}
\usepackage[numbers,sort&compress]{natbib}
\usepackage[colorlinks=true,citecolor=blue!60!black,linkcolor=blue!60!black,urlcolor=blue!60!black]{hyperref}
\usepackage[capitalise,noabbrev]{cleveref}

\newcommand{\ours}{\textsc{SelfWrite}}

\newcommand{\Mg}{M_g}
\newcommand{\Mz}{M_0}

\newcommand{\gate}{\mathcal{G}}

\title{Recursive Self-Editing: LLMs Program Their Own Weight Edits for Self-Improvement}
\author{Linghua Zhang \and Yunhe Li}
\hypersetup{pdfauthor={Linghua Zhang, Yunhe Li}}
\date{}

\begin{document}
\maketitle
\begin{abstract}
Recursive self-improvement asks a system to improve the very system that improves it.
Today's language models improve themselves through data, training recipes or scaffolds; whenever their weights change, gradient descent writes the update, not the model.
We let the model write its own weights.
In \ours{}, each generation the current model studies its own reasoning on fresh training problems, inspects its own weights and activations in a persistent Python session, and writes a program that constructs a low-rank weight edit for itself from white-box primitives (singular directions, activation contrasts between its right and wrong answers, rank-one projections, tensor computations), with no training or gradient tool.
Every candidate is graded once on a hidden gate set against the current model; the best strictly improving edit is merged into the weights, and the edited model writes the next edit.
On Qwen3.8-27B and math problems from DAPO-Math, five generations of \ours{} raise gate accuracy from 73.95\% to 76.45\% and accuracy on a held-out test split that the loop never sees from 76.30\% to 77.75\% (avg@4) and from 82.60\% to 84.80\% (pass@4), without a single gradient step.
A language model can author useful changes to its own weights without gradients, and the edited model keeps writing useful edits.

\end{abstract}

\section{Introduction}\label{sec:intro}

A recursively self-improving system improves the very system that does the improving \citep{good1965,godelmachine}.
Large language models now improve themselves along three routes: they generate and judge their own training \emph{data} \citep{selfrewarding,rzero,absolutezero}, choose \emph{training recipes} for themselves or other models \citep{seal,metarsi,ftdojo,posttrainbench}, or rewrite the \emph{scaffolds and harnesses} around a model \citep{stop,godelagent,dgm,aide2,hase,selfhpo}.
In SEAL \citep{seal}, for example, the model writes synthetic data and hyper-parameters, and LoRA fine-tuning \citep{lora} turns them into a weight change: the model decides what to train on, and SGD decides how every parameter moves.

\begin{figure*}[t]
  \centering
\resizebox{\textwidth}{!}{%
\begin{tikzpicture}[
  font=\small,
  >={Stealth[length=2.4mm,width=2.0mm]},
  box/.style={draw=#1!85!black, fill=#1!10, rounded corners=2pt, align=center, inner sep=3.5pt,
              line width=0.7pt, minimum height=1.15cm},
  flow/.style={->, line width=0.9pt, draw=black!75},
  band/.style={fill=#1!6, draw=#1!40, rounded corners=4pt, line width=0.5pt},
  bandtitle/.style={anchor=north west, font=\normalsize\bfseries, text=#1!80!black},
  llm/.pic={
    \draw[rounded corners=3pt, line width=0.8pt, draw=black!70, fill=white] (-0.62,-0.70) rectangle (0.62,0.70);
    \foreach \y in {0.42,0.14,-0.14,-0.42}
      \fill[#1, rounded corners=1pt] (-0.44,\y-0.09) rectangle (0.44,\y+0.09);
  },
]
\definecolor{dteal}{HTML}{1B9E77}
\definecolor{dorange}{HTML}{D95F02}
\definecolor{dgrey}{HTML}{666666}
\definecolor{dblue}{HTML}{5E7A8C}
\definecolor{dred}{HTML}{C0392B}
\def\code#1{{\ttfamily\fontsize{8.6}{10.2}\selectfont #1}}

\begin{scope}[on background layer]
  \draw[band=dgrey] (-0.55, 3.15) rectangle (17.55, 5.75);
  \draw[band=dteal] (-0.55,-1.55) rectangle (17.55, 2.95);
\end{scope}
\node[bandtitle=dgrey] at (-0.45,5.70) {Existing: self-improvement through data or training recipes};
\node[bandtitle=dteal] at (-0.45,2.90) {\textsc{SelfWrite} (ours): the model writes its own weights};

\pic at (0.85,4.25) {llm=dblue!55};
\node[font=\footnotesize] at (0.85,3.36) {model $M$};
\node[box=dgrey, text width=3.0cm] (data) at (3.9,4.25)
  {\textbf{writes data or picks}\\[-1pt]\textbf{hyper-parameters}};
\node[box=dorange, text width=4.2cm, line width=1.6pt, fill=dorange!16] (sgd) at (8.25,4.25)
  {{\Large\bfseries SGD}\\[1pt] $\theta \leftarrow \theta - \eta\,\nabla_{\theta}\mathcal{L}$};
\pic at (11.55,4.25) {llm=dorange!55};
\node[font=\footnotesize] at (11.55,3.36) {model $M'$};
\node[box=dred, fill=dred!6, text width=4.0cm] (tag1) at (15.15,4.25)
  {\textcolor{dred}{\bfseries\itshape the gradient writes the weights}\\[1pt]
   {\footnotesize the model only chooses the inputs}};
\draw[flow] (1.55,4.25) -- (data.west);
\draw[flow] (data.east) -- (sgd.west);
\draw[flow] (sgd.east) -- (10.85,4.25);
\draw[flow] (12.25,4.25) -- (tag1.west);

\pic at (0.85,0.95) {llm=dteal!60};
\node[font=\footnotesize, align=center] at (0.85,-0.05) {$M_g$ = editor\\[-2pt] = edited model};
\node[box=dteal, text width=3.55cm, align=left] (ans) at (3.95,0.95)
  {\textbf{reads its own answers}\\[1pt]
   {\footnotesize\textcolor{dred}{$\boldsymbol{\times}$}\ \#217:\ 41 \ (gold 45)}\\[-1pt]
   {\footnotesize\textcolor{dteal!80!black}{$\boldsymbol{\checkmark}$}\ \#032:\ 12 \ (gold 12)}\\[-1pt]
   {\footnotesize\itshape with full reasoning traces}};
\node[box=dteal, text width=4.95cm, align=left] (prog) at (8.85,0.95)
  {\textbf{writes a program}\\[1pt]
   \code{w = acts(m, wrong, "own")}\\
   \code{r = acts(m, right, "own")}\\
   \code{u = direction(w, r)}\\
   \code{return project\_out(m, u)}};
\begin{scope}[shift={(12.35,0.95)}]
  \fill[dteal!70] (-0.55,-0.50) rectangle (-0.40,0.50);
  \node[font=\footnotesize] at (-0.18,0.05) {$\times$};
  \fill[dteal!70] (0.02,0.35) rectangle (0.72,0.50);
  \node[font=\footnotesize, align=center] at (0.08,-0.82) {rank-1 $\Delta W$};
\end{scope}
\node[box=dorange, text width=2.2cm] (gate) at (14.75,0.95)
  {\textbf{hidden gate}\\[-1pt]{\footnotesize 500 problems}\\[-1pt]
   $\Delta_{\mathcal{G}} > 0$\ \textcolor{dteal!80!black}{$\boldsymbol{\checkmark}$}};
\pic at (16.85,0.95) {llm=dteal!60};
\node[font=\footnotesize, align=center] at (16.85,-0.05) {$M_{g+1}$\\[-2pt] merged};
\draw[flow] (1.55,0.95) -- (ans.west);
\draw[flow] (ans.east) -- (prog.west);
\draw[flow] (prog.east) -- (11.70,0.95);
\draw[flow] (13.20,0.95) -- (gate.west);
\draw[flow] (gate.east) -- (16.15,0.95);
\begin{scope}[shift={(11.75,2.30)}]
  \draw[dgrey, line width=0.8pt, fill=white] (0,0) circle (0.27);
  \node at (0,-0.01) {$\nabla$};
  \draw[dred, line width=1.6pt] (-0.19,-0.19) -- (0.19,0.19);
  \node[anchor=west, font=\footnotesize, text=dred!85!black] at (0.33,0) {no gradient step, no training};
\end{scope}
\draw[->, line width=1.6pt, draw=dteal!85!black, rounded corners=8pt]
  (16.85,-0.38) -- (16.85,-1.15) -- (0.85,-1.15) -- (0.85,-0.42);
\node[font=\footnotesize\bfseries, text=dteal!75!black, fill=dteal!6, inner sep=2pt] at (8.7,-1.15)
  {the edited model $M_{g+1}$ writes the next edit};
\end{tikzpicture}%
}
  \caption{\textbf{The gradient writes the weights, or the model does.}
  Top: in existing self-improvement the model writes data or chooses hyper-parameters, and SGD computes the weight change.
  Bottom: in \ours{} the model reads its own wrong and right answers, contrasts its activations on them, and writes a program that projects the contrast direction out of one of its layers (code simplified).
  The rank-one edit is merged only if it improves the model on a hidden gate set, and the edited model $M_{g+1}$ writes the next edit. No gradient tool is used and no training step is taken.}
  \label{fig:teaser}
\end{figure*}
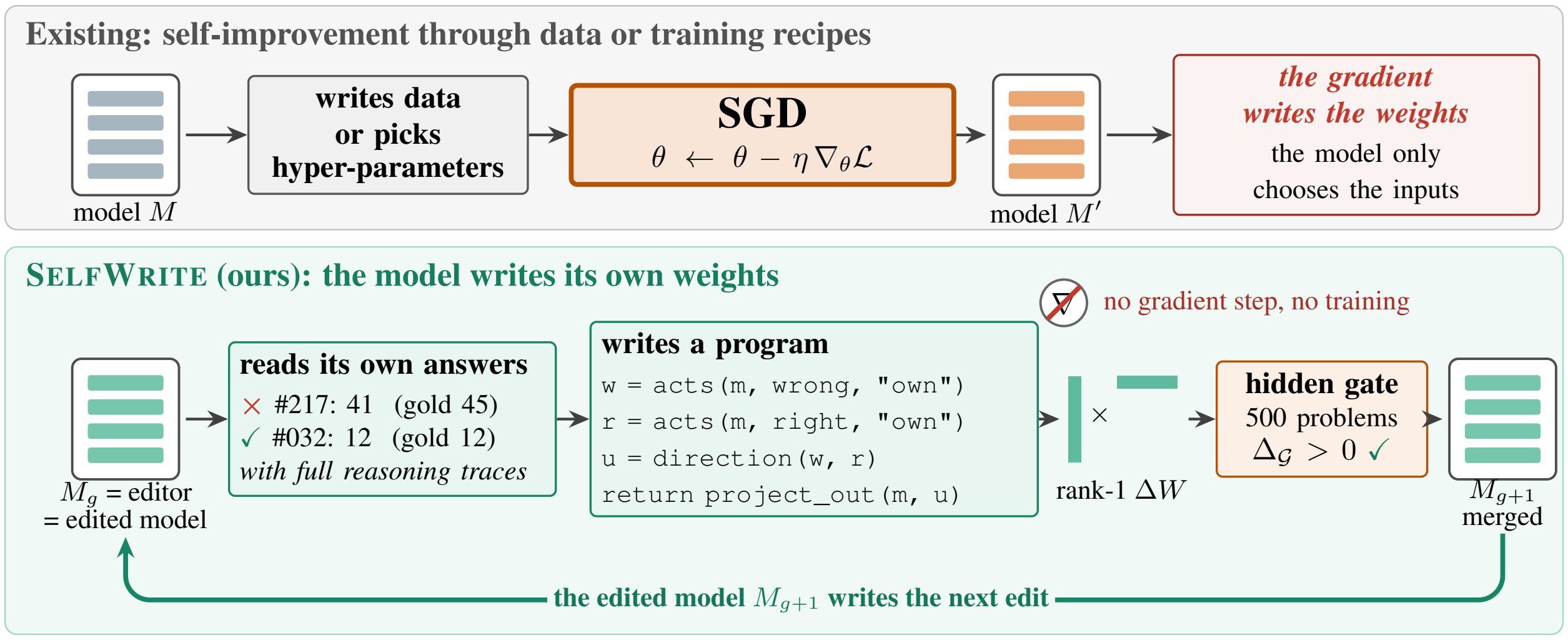

We take the other route: the model writes its own weights.
A model that has just answered a batch of math problems reads its own outputs, contrasts the activations of one of its layers on the problems it failed and the ones it solved, and removes the contrast direction from the layer's output matrix with a rank-one edit.
If the edited model solves more held-out problems, the edit is merged, and the edited model, now reading its own changed weights, writes the next edit.
No tool computes a gradient and no training step is taken (\cref{fig:teaser}).

This route addresses two gaps in prior work.
First, \emph{the gradient writes the weights.}
In data- and recipe-level self-improvement the model chooses the inputs of an optimiser \citep{seal,selfspecialization,mass,metarsi}; methods that emit weights directly rely on generators trained for that purpose \citep{generativeadapter,text2lora,transformer2}; and gradient-free weight programs are written by an external model for a different network \citep{wsp} or applied to a fixed base model \citep{selfdeveloping}.
Second, \emph{the improved model rarely improves its own weights next.}
SEAL discards the adapted weights after each episode, MetaRSI retrains from the base model \citep{seal,metarsi}, and where the updated model does take over, its weights are still changed by fine-tuning \citep{mass} or it edits harnesses rather than weights \citep{hase,selfhpo}.

We formulate \emph{gradient-free recursive self-editing}: in generation $g$ the current model $\Mg$ writes a program that constructs a weight edit $\Delta_g$ for itself; if $\Delta_g$ improves $\Mg$ on a gate set the editor never sees, it is merged, $M_{g+1} = \Mg \oplus \Delta_g$, and $M_{g+1}$ writes the next edit.
The editor has no gradient tool and no stronger external model, grading is deterministic, and edits are capped in rank and norm.
The setting is demanding: among hundreds of weight matrices the editor must find the few directions that help reasoning, every candidate must be graded at 27B scale on long sampled reasoning traces, and the signal the editor reads must be kept apart from the decision that accepts an edit.

\begin{figure*}[t]
  \centering
\resizebox{\textwidth}{!}{%
\begin{tikzpicture}[
  font=\small,
  >={Stealth[length=2.4mm,width=2.0mm]},
  panel/.style={draw=#1!55, fill=#1!6, rounded corners=5pt, line width=0.8pt},
  ptitle/.style={anchor=north west, font=\normalsize\bfseries, text=#1!80!black},
  num/.style={circle, fill=#1!85!black, text=white, font=\small\bfseries, inner sep=1.4pt, minimum size=0.48cm},
  box/.style={draw=#1!80!black, fill=white, rounded corners=2pt, align=center, inner sep=3pt, line width=0.6pt},
  flow/.style={->, line width=1.0pt, draw=black!75},
  llm/.pic={
    \draw[rounded corners=3pt, line width=0.8pt, draw=black!70, fill=white] (-0.55,-0.62) rectangle (0.55,0.62);
    \foreach \y in {0.37,0.12,-0.13,-0.38}
      \fill[#1, rounded corners=1pt] (-0.39,\y-0.08) rectangle (0.39,\y+0.08);
  },
  card/.style={draw=dteal!80!black, fill=white, rounded corners=2pt, line width=0.6pt,
               minimum width=1.30cm, minimum height=1.05cm},
]
\definecolor{dteal}{HTML}{1B9E77}
\definecolor{dorange}{HTML}{D95F02}
\definecolor{dgrey}{HTML}{666666}
\definecolor{dblue}{HTML}{5E7A8C}
\definecolor{dred}{HTML}{C0392B}
\def\code#1{{\ttfamily\fontsize{8.2}{9.6}\selectfont #1}}

\begin{scope}[on background layer]
  \draw[panel=dblue]   (0.00,0.00) rectangle (4.40,5.40);
  \draw[panel=dteal]   (5.00,0.00) rectangle (11.85,5.40);
  \draw[panel=dorange] (12.45,0.00) rectangle (17.40,5.40);
\end{scope}
\node[num=dblue]   at (0.40,5.00) {1}; \node[ptitle=dblue]   at (0.70,5.27) {Observe};
\node[num=dteal]   at (5.40,5.00) {2}; \node[ptitle=dteal]   at (5.70,5.27) {Write: 8 independent sessions};
\node[num=dorange] at (12.85,5.00) {3}; \node[ptitle=dorange] at (13.15,5.27) {Select: hidden gate};

\pic at (0.95,4.05) {llm=dteal!60};
\node[font=\small, anchor=west, align=left] at (1.62,4.05) {$M_g$ answers 500\\[-1pt] fresh problems,\\[-1pt] 4 samples each};
\node[box=dblue, text width=3.55cm, align=left] (sheet) at (2.20,1.85)
  {\textbf{reads its own answers}\\[0pt]
   {\footnotesize\textcolor{dred}{$\boldsymbol{\times}$}\,wrong\ \ \textcolor{dteal!80!black}{$\boldsymbol{\checkmark}$}\,right}\\[-1pt]
   {\footnotesize\itshape with full reasoning traces}\\[2pt]
   \code{screen\_results}\\ \code{screen\_output}\\ \code{acts(source="own")}};

\node[font=\small, align=center] at (8.42,4.40)
  {editor $= M_g$ (thinking), $\le$30 tool rounds\\[-1pt] in a persistent Python session};
\foreach \k in {1,...,8}{
  \pgfmathtruncatemacro{\row}{(\k-1)/4}
  \pgfmathsetmacro{\cx}{5.95 + mod(\k-1,4)*1.62}
  \pgfmathsetmacro{\cy}{3.20 - \row*1.22}
  \node[card] (c\k) at (\cx,\cy) {};
  \node[font=\footnotesize, anchor=north west] at ($(c\k.north west)+(0.04,-0.02)$) {s\k};
  \fill[dteal!65] ($(c\k.center)+(-0.42,-0.38)$) rectangle ($(c\k.center)+(-0.30,0.10)$);
  \fill[dteal!65] ($(c\k.center)+(-0.22,-0.02)$) rectangle ($(c\k.center)+(0.42,0.10)$);
  \node[font=\scriptsize] at ($(c\k.center)+(0.14,-0.24)$) {$\Delta W_{\k}$};
}
\node[box=dteal, fill=dteal!8, minimum height=0.90cm, text width=5.05cm] (kit) at (7.92,0.68)
  {\textbf{white-box toolkit}\\[0pt]
   {\footnotesize read $\cdot$ construct $\cdot$ compute $\cdot$ edit ops}};
\begin{scope}[shift={(11.15,1.10)}]
  \draw[dgrey, line width=0.7pt, fill=white] (0,0) circle (0.24);
  \node[font=\small] at (0,-0.01) {$\nabla$};
  \draw[dred, line width=1.5pt] (-0.17,-0.17) -- (0.17,0.17);
\end{scope}
\node[font=\scriptsize, text=dred!85!black, anchor=north, align=center] at (11.15,0.83) {no\\gradient};

\begin{scope}[shift={(13.20,2.95)}]
  \draw[line width=0.6pt, black!60] (-0.10,0) -- (4.05,0);
  \node[font=\footnotesize, anchor=east] at (-0.12,0) {0};
  \foreach \k/\h in {1/-0.35,2/0.55,3/-0.20,4/1.10,5/0.20,6/-0.62,7/0.45,8/0.10}{
    \pgfmathsetmacro{\x}{(\k-1)*0.50+0.05}
    \ifnum\k=4
      \fill[dteal] (\x,0) rectangle (\x+0.34,\h);
    \else
      \fill[black!28] (\x,0) rectangle (\x+0.34,\h);
    \fi
    \node[font=\scriptsize, anchor=north] at (\x+0.17,-0.68) {s\k};
  }
  \node[font=\footnotesize, text=dteal!80!black, anchor=south] at (1.72,1.12) {best $\Delta_{\mathcal G}>0$};
  \node[font=\footnotesize, anchor=north] at (1.95,-1.00) {paired gain $\Delta_{\mathcal G}$ vs.\ $M_g$, 500 problems};
\end{scope}
\node[box=dteal, line width=1.0pt, fill=dteal!10, text width=4.3cm] (merge) at (14.92,0.62)
  {\textbf{merge the best edit}\\[-1pt] $W' = W + BA \;\Rightarrow\; M_{g+1}$};

\node[draw=dgrey, dashed, fill=black!3, rounded corners=2pt, font=\footnotesize, align=center, inner sep=3pt]
  (test) at (16.55,-0.80) {test (500):\\[-1pt] never shown to the editor};
\draw[->, dgrey, dashed, line width=0.8pt] (16.55,0.00) -- (test.north);

\draw[flow] (4.40,2.70) -- (5.00,2.70);
\draw[flow] (11.85,2.70) -- (12.45,2.70);
\draw[->, dorange!85!black, dashed, line width=0.8pt, rounded corners=5pt]
  (14.92,5.40) -- (14.92,5.90) -- (8.42,5.90) -- (8.42,5.40);
\node[font=\footnotesize, text=dorange!85!black, fill=white, inner sep=1.5pt] at (11.65,5.90)
  {gate statistics (no problem content) $\to$ sessions of later generations};
\draw[->, line width=1.8pt, draw=dteal!85!black, rounded corners=9pt]
  (14.92,0.00) -- (14.92,-0.60) -- (2.20,-0.60) -- (2.20,0.00);
\node[font=\small\bfseries, text=dteal!75!black, fill=white, inner sep=2pt] at (8.56,-0.60)
  {succession: $M_{g+1}$ is the next editor (new screen, same gate)};
\end{tikzpicture}%
}
  \caption{\textbf{One generation of \ours{}.}
  \textbf{(1) Observe:} the current model $\Mg$ answers 500 freshly drawn problems four times, and the editor reads its own answers, reasoning included (\cref{sec:method-observe}).
  \textbf{(2) Write:} eight independent sessions of the editor $\Mg$, each with up to 30 tool rounds over a white-box toolkit for its own weights, each return a program that builds one LoRA $\Delta W_k$ (\cref{sec:method-primitives}).
  \textbf{(3) Select:} every candidate is graded once on the hidden gate, paired problem by problem with $\Mg$; the largest strictly positive gain is merged, and $M_{g+1}$ becomes the next editor (\cref{sec:method-gate}); gate statistics reach the sessions of later generations. Bars are schematic. The test split is never shown to the editor.}
  \label{fig:overview}
\end{figure*}
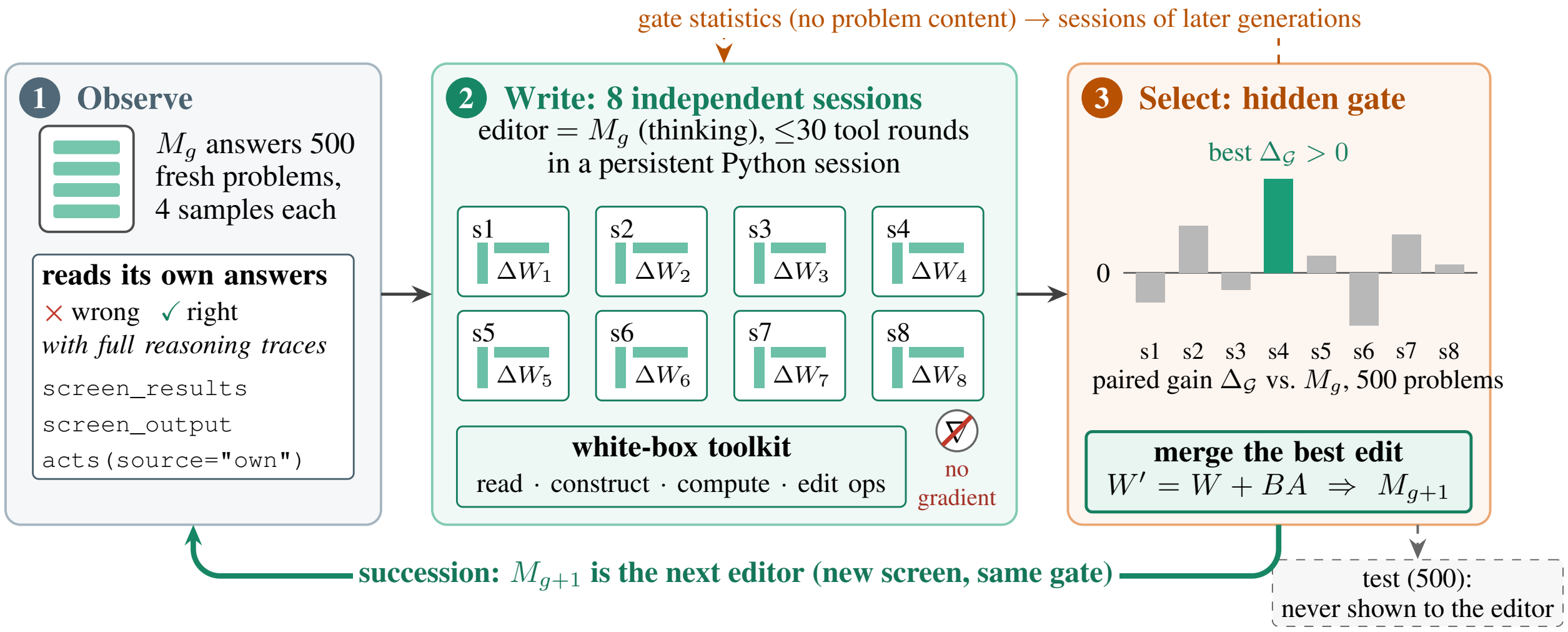

\ours{} meets these requirements with four components (\cref{fig:overview}).
\textbf{(i) Observe.} Each generation, $\Mg$ answers 500 freshly drawn training problems four times in thinking mode, and the editor can read every one of its own answers, including the full reasoning, and compute activation statistics on them.
\textbf{(ii) Write.} The editor works in a persistent Python session for up to 30 tool rounds, with a white-box toolkit over its own weights: truncated SVD of any module, activation statistics on its own right and wrong answers, rank-one constructions and projections, singular-direction rescaling, LoRA operators, and sandboxed tensor computations on any weight; it then returns a self-contained program that builds the edit.
\textbf{(iii) Select on a hidden paired gate.} Eight independent sessions per generation each produce one candidate; every candidate is graded once on a fixed 500-problem gate set, paired item by item with $\Mg$, and later generations see only the statistics of earlier gate results, never a gate problem's statement or answer.
\textbf{(iv) Merge and succeed.} The candidate with the largest strictly positive paired gain is merged into the weights, and the merged model becomes the next editor, with a searchable history of every earlier attempt.

We study \ours{} on Qwen3.8-27B \citep{qwen38} for five generations on DAPO-Math-17K \citep{dapo}, tracking gate accuracy after every generation and reporting a held-out test split that is never screened, gated or shown to the editor.
Our contributions are:
\begin{itemize}
  \item \textbf{Setting.} We formulate gradient-free recursive self-editing, in which the current model writes the program that constructs its own weight edit and the merged model edits next, under a protocol with no gradient tool, disjoint screen, gate and test data, and deterministic grading (\cref{sec:method-setting}).
  \item \textbf{Method.} \ours{} combines self-observation of the model's own reasoning, an agentic editing session with a white-box weight toolkit, a hidden paired gate, and a merge-and-succeed loop, made affordable at 27B by hot-loading candidate LoRAs on data-parallel replicas (\cref{sec:method}).
  \item \textbf{Experiments.} On math, five generations of \ours{} raise gate accuracy from 73.95\% to 76.45\% and held-out test accuracy from 76.30\% to 77.75\% (avg@4) and from 82.60\% to 84.80\% (pass@4), with every gain produced by an edit the model wrote for itself (\cref{sec:results}).
\end{itemize}

\section{Related Work}\label{sec:related}

\paragraph{Recursive self-improvement.}
\citet{good1965} imagined machines that improve the machinery that improves them, and the Gödel machine~\citep{godelmachine} accepts a self-rewrite once it is proven useful.
Recent LLM systems replace the proof with an empirical check and act on code: STOP~\citep{stop} improves a scaffold with a fixed model, Gödel Agent~\citep{godelagent} rewrites its runtime logic, the Darwin Gödel Machine~\citep{dgm} evolves coding agents around a frozen foundation model, and AIDE$^2$~\citep{aide2} accepts agent rewrites that win on private held-out tasks.
Closer to weights, MASS~\citep{mass} merges a LoRA trained on a model's self-evaluated trajectories, SIA~\citep{sia} lets an external agent choose between harness edits and LoRA reinforcement-learning updates, and MetaRSI~\citep{metarsi} proposes LoRA training recipes from a bounded space.
In all of them a fixed optimiser makes any weight change; in \ours{} the model writes the change itself, and the edited model is the next editor.
Because recursive self-training can drift over rounds \citep{recursivecollapse}, we follow AIDE$^2$ and accept an edit only on a fixed external check whose problems the editor never sees.

\paragraph{Self-improvement through data and recipes.}
Models produce their own training signal by judging their responses~\citep{selfrewarding}, proposing and solving their own tasks~\citep{absolutezero,rzero}, training experts on self-generated instructions~\citep{selfspecialization}, or fitting temporary adapters at test time~\citep{ttsi}.
SEAL~\citep{seal} writes self-edits (synthetic data and optimisation settings) that a fixed fine-tuning step applies, and searching over self-edit strategies extends this to templates and LoRA settings~\citep{searchselfedit}.
Agents also run post-training pipelines~\citep{ftdojo,posttrainbench}, write merging algorithms~\citep{selfdeveloping}, or alternate harness edits with fine-tuning~\citep{hase,whale,selfhpo}.
In each case the model chooses the inputs of an optimiser and gradient descent writes the weights; in \ours{} the program the model writes \emph{is} the weight edit.

\paragraph{Generated weights, LoRA editing and weight-space programming.}
Generative Adapter~\citep{generativeadapter} and Text-to-LoRA~\citep{text2lora} train a generator that maps a context to low-rank weights; Transformer$^2$~\citep{transformer2} learns singular-value scaling vectors, Self-MoE~\citep{selfmoe} routes among self-specialised experts, and SOLAR~\citep{solar} trains a policy that proposes weight-adaptation strategies.
LoRA~\citep{lora} and PiSSA~\citep{pissa} write updates as low-rank products, intruder directions characterise how LoRA changes the spectrum~\citep{intruder}, and adapters are combined in closed form~\citep{taskarith,ties,dare,regmean}, by searching merge coefficients~\citep{lorahub,evomerge,evogm,auditmerge}, or edited by fixed rules~\citep{para,tada,slora}; ROME and MEMIT write closed-form rank-one factual edits~\citep{rome,memit}.
Weight-space programming lets an LLM write code that edits network weights~\citep{wsp,wspscope,wsprobots,weightloop}.
\ours{} builds on these operators but changes who uses them and on what: the edited model itself composes them, guided by its own reasoning traces, and each accepted edit changes the model that writes the next one.

\section{Method}\label{sec:method}

\subsection{Setting}\label{sec:method-setting}
Let $\Mz$ be the original model and $\Mg$ the model after $g$ accepted generations.
In generation $g$, $\Mg$ acts as the editor and writes programs whose output is a LoRA $\Delta_g = \{(A_m, B_m)\}_{m\in\mathcal{M}_g}$ over a set of linear modules $\mathcal{M}_g$.
A candidate must pass the checks
\begin{equation}
  \operatorname{rank}(B_m A_m) \le 64, \qquad \|B_m A_m\|_F \,/\, \|W_g^{(m)}\|_F \le \tau = 0.1 \quad \forall m \in \mathcal{M}_g,
  \label{eq:caps}
\end{equation}
where $W_g^{(m)}$ is the current weight of module $m$, and returning no edit (keeping $\Mg$) is always allowed.
An accepted edit is merged,
\begin{equation}
  W_{g+1}^{(m)} = \mathrm{bf16}\big(W_g^{(m)} + B_m A_m\big), \qquad m \in \mathcal{M}_g,
  \label{eq:merge}
\end{equation}
with accumulation in FP32, and the merged model $M_{g+1}$ is the editor of generation $g+1$.

\paragraph{Data roles.}
All problems come from one decontaminated training pool, split once into three disjoint parts: a \emph{gate} set $\gate$ of 500 problems that decides acceptance, a \emph{test} set of 500 problems that is reported and never fed back, and the rest.
Each generation draws a fresh \emph{screen} set $S_g$ of 500 problems from the rest; the remaining problems form the generation's \emph{train} set, whose statements and gold answers programs may read.
Scoring samples four answers per problem in thinking mode (temperature 0.6, top-$p$ 0.95, top-$k$ 20, up to 16,384 tokens, fixed seed), and a problem's score $s(x)\in[0,1]$ is the fraction of correct answers under deterministic answer matching.
For a candidate $c$ and the current model $\Mg$, the paired gain on $\gate$ is
\begin{equation}
  \Delta_{\gate}(c) = \frac{1}{|\gate|}\sum_{x\in\gate}\big(s_c(x) - s_{\Mg}(x)\big),
  \label{eq:paired-b}
\end{equation}
which editors of later generations receive as feedback together with a 95\% paired-bootstrap confidence interval and the counts and ids of the problems fixed and broken.

\subsection{Observe: the model reads its own answers}\label{sec:method-observe}
At the start of each generation $\Mg$ answers the 500 screen problems once (four samples each).
Every editing message carries a summary of this run: accuracy, the distribution of correct samples per problem, the truncation rate, and a few failed problems with statement, gold answer and extracted answer.
Two free tools expose the full record: \texttt{screen\_results} filters problems by outcome (all; wrong, unstable or right: 0, 1--3 or 4 of 4 samples correct), and \texttt{screen\_output} pages through any sampled answer including its reasoning.
The same answers feed activation statistics: \texttt{acts(source='own')} runs the current weights on the problem \emph{plus the model's own answer} (its first sample), so the editor can contrast how a layer behaves when it reasons correctly and when it fails.

\subsection{Write: agentic editing sessions}\label{sec:method-primitives}
Each candidate is written in one session of the editor $\Mg$ in thinking mode.
The session offers three chat tools: \texttt{run\_python} executes code in a sandboxed Python process that persists across calls (variables, handles and imports survive), \texttt{read\_record} opens the program and feedback of any earlier candidate, and \texttt{search\_history} lists every candidate tried in this and all earlier generations with its modules, rank, tools used and gate outcome.
Inside \texttt{run\_python} the editor calls the toolkit in \cref{tab:primitives} through an \texttt{api} object.
After at most 30 tool rounds the editor returns a self-contained program \texttt{main(api)} that rebuilds the edit; the program runs in a fresh sandbox and its LoRA is checked against \cref{eq:caps}.
Intermediate edits can be stored with \texttt{save\_edit} and reloaded by the final program.

\begin{table}[t]
\centering
\footnotesize
\setlength{\tabcolsep}{4pt}
\caption{The editor's toolkit (full signatures in \cref{tab:app-tools}). $W$: current weight of a module; $(u_i,\sigma_i,v_i)$: its singular triplets. No tool computes a gradient or trains.}
\label{tab:primitives}
\begin{tabular}{@{}lp{0.84\linewidth}@{}}
\toprule
Observe & \texttt{screen\_results}, \texttt{screen\_output} (own answers, reasoning included), \texttt{train\_example}, \texttt{diagnostics} \\
Read & \texttt{svd} (top-$k$ triplets of $W$), \texttt{acts} (activation statistics on own answers or prompts), \texttt{tensors}, \texttt{output\_matrices} \\
Construct & \texttt{rank1} ($s\,uv^\top$), \texttt{from\_direction} ($\Delta W k = \alpha d$), \texttt{scale\_singular} ($\sum_i (f_i{-}1)\sigma_i u_i v_i^\top$), \texttt{project\_out} ($-\alpha\, uu^\top W$), \texttt{direction}, \texttt{unit} \\
Compute & \texttt{tensor\_compute}: sandboxed CPU PyTorch on up to four FP32 weight tensors $\to$ statistics, vectors or LoRA factors \\
Operate & \texttt{scale}, \texttt{interpolate}, \texttt{combine}, \texttt{drop}, \texttt{truncate\_rank}, \texttt{intruder\_scale}, \texttt{orthogonalize}, \texttt{ties}, \dots \\
Check & \texttt{inspect}, \texttt{train\_loss} (forward-only NLL on train problems), \texttt{compare} (fixed / broken problems) \\
\bottomrule
\end{tabular}
\end{table}

The toolkit supports edits that are grounded in the model's own behaviour.
A typical construction, suggested in the editor's prompt, collects activations of one layer on the model's own correct and incorrect answers, forms the normalised difference $d = (\bar a_{\text{wrong}} - \bar a_{\text{right}})/\|\bar a_{\text{wrong}} - \bar a_{\text{right}}\|$, and removes it from the layer's output matrices with \texttt{project\_out}, which adds the rank-one term $B = -\alpha d$, $A = d^\top W$.
Other programs rescale a few singular directions of selected modules, add activation directions with \texttt{from\_direction}, or compute exact factors with \texttt{tensor\_compute}; \texttt{train\_loss} provides a cheap forward-only check on train problems before a candidate is graded.
Activation statistics and losses are charged to a GPU-time budget, while observation, reading and construction are free.
\Cref{fig:program-b} shows a program of this kind.

\begin{figure}[t]
\centering
\begin{minipage}{0.92\linewidth}
\footnotesize
\begin{verbatim}
def main(api):
    wrong = [r["id"] for r in api.screen_results(filter="wrong", limit=64)]
    right = [r["id"] for r in api.screen_results(filter="right", limit=64)]
    lora = api.new_lora()["lora"]
    for layer in (44, 48, 52):
        mod = f"layers.{layer}.mlp.down_proj"
        a = api.acts(module=mod, ids=wrong, source="own")
        b = api.acts(module=mod, ids=right, source="own")
        d = api.direction(a=a["handle"] + ".out_mean",
                          b=b["handle"] + ".out_mean")
        for m in api.output_matrices(layer=layer):
            lora = api.project_out(target=lora, module=m["module"],
                                   u=d["vector"], alpha=0.5)["lora"]
    # forward-only sanity check on train problems
    if api.train_loss(lora=lora, n=64)["delta"] > 0:
        lora = api.scale(lora=lora, s=0.5)["lora"]
    return {"lora": lora}
\end{verbatim}
\end{minipage}
\caption{An illustrative \ours{} program (real tool names and signatures; layers and constants are examples): contrast the model's activations on its own wrong and right answers, and project the contrast direction out of the layers' output matrices.}
\label{fig:program-b}
\end{figure}

\subsection{Select: hidden gate, merge and succession}\label{sec:method-gate}
\begin{algorithm}[t]
\caption{One generation of \ours{}.}
\label{alg:selfwrite}
\begin{algorithmic}[1]
\Require current model $\Mg$; gate set $\gate$; test set; clean pool $P$; archive and history $\mathcal{A}$
\State $S_g \gets$ 500 problems drawn afresh from $P \setminus (\gate \cup \text{test})$; train set $T_g \gets P \setminus (\gate \cup \text{test} \cup S_g)$
\State $\Mg$ answers $S_g$ (4 samples, thinking mode); store answers, reasoning and summary
\State score $\Mg$ on $\gate$ (4 samples, same decoding and seed)
\For{$k = 1, \dots, 8$} \Comment{independent sessions, run in parallel}
  \State editor $\Mg$ explores with \texttt{run\_python}, \texttt{read\_record}, \texttt{search\_history} ($\le 30$ tool rounds)
  \State editor returns program $p_k$; run $p_k$ in a fresh sandbox $\Rightarrow$ LoRA $c_k$
  \If{$c_k$ passes the caps (\cref{eq:caps})}
    \State hot-load $c_k$ on all replicas; score on $\gate$; $\Delta_k \gets \Delta_{\gate}(c_k)$ (\cref{eq:paired-b})
    \State record $\Delta_k$, its CI and fixed / broken counts and ids for later generations
  \EndIf
\EndFor
\State $k^\star \gets \arg\max_k \Delta_k$ over candidates with $\Delta_k > 0$
\If{$k^\star$ exists} \State $M_{g+1} \gets \Mg \oplus c_{k^\star}$ (\cref{eq:merge}) \Else \State $M_{g+1} \gets \Mg$ \EndIf
\State add the generation's programs and outcomes to $\mathcal{A}$; score $M_{g+1}$ on the test set (logged, never shown to the editor)
\end{algorithmic}
\end{algorithm}

A generation runs eight independent sessions in parallel, each producing one candidate. \Cref{alg:selfwrite} summarises one generation.
Every candidate is graded once on $\gate$ (500 problems, four samples each) and compared with $\Mg$ by \cref{eq:paired-b}; the scores of $\Mg$ on $\gate$ are computed once per generation with the same decoding and seed.
The editor never reads a gate problem: sessions of later generations see only each earlier candidate's accuracy, paired gain, confidence interval and the counts and ids of fixed and broken problems, and no tool accepts a gate id.
At the end of the generation, the candidate with the largest strictly positive paired gain is accepted and merged by \cref{eq:merge}; if none improves, $\Mg$ is kept.
The loop stops after three consecutive generations without an accepted edit. After each generation the current model is also scored on the test split for logging; these scores are never shown to the editor, and we report $\Mz$ and the final model.
Earlier generations' best programs are shown to the next editor and every attempt stays searchable, so experience accumulates while the editing model itself changes.

\paragraph{Safe self-modification.}
The loop changes weights only through programs that run in a sandbox without network access, read only train and screen problems, and produce edits that pass the caps in \cref{eq:caps}.
Acceptance is decided by deterministic grading on data the editor cannot read, never by the model's own judgement, and every accepted edit is a stored program plus a low-rank tensor that a human can audit or revert.

\subsection{Implementation}\label{sec:method-impl}
The editor writes each LoRA from zero; nothing is trained at any point.
Seven data-parallel vLLM \citep{vllm} replicas of $\Mg$ serve the editor, hot-load candidate LoRAs and split the gate problems, and one GPU holds a Transformers copy of $\Mg$ for SVD, activation statistics, forward losses and merging (\cref{sec:app-method}).

\section{Experimental Setup}\label{sec:exp}

\paragraph{Model and data.}
We use Qwen3.8-27B \citep{qwen38}, a dense model with full-attention and Gated DeltaNet layers, as $\Mz$; the same model is the first editor.
All problems come from DAPO-Math-17K \citep{dapo} after decontamination against MATH, GSM8K, AIME, AMC and OlympiadBench (exact match after normalisation, or at least half of the 13-grams shared) and removal of internal duplicates, which leaves 14,738 problems.
From this pool we fix a gate set of 500 problems and a test set of 500 problems, pairwise disjoint by id and by statement.
Each generation draws a fresh screen set of 500 problems from the rest; the remaining 13,238 problems are that generation's train set.

\paragraph{Protocol.}
We run five generations of \ours{}.
Each generation has eight independent editing sessions with up to 30 tool rounds each; tool turns and the final turn are capped at 16,384 output tokens, and the editor samples in thinking mode (temperature 0.6, top-$p$ 0.95, top-$k$ 20).
Screen, gate and test scoring use the same thinking-mode decoding with four samples per problem (avg@4) and a fixed seed.
The candidate with the largest strictly positive paired gain on the gate set is merged.
Runs use one node with eight GPUs: seven vLLM \citep{vllm} replicas of $\Mg$ and one tool GPU (\cref{sec:method-impl}); all settings are listed in \cref{tab:hparams}.

\paragraph{Metrics.}
We report the gate accuracy (avg@4) of the current model after every generation, the paired gain $\Delta_{\gate}$ of each accepted edit over the model that wrote it (\cref{eq:paired-b}), and the test accuracy of $\Mz$ and of the final model on the 500 test problems, both as avg@4 (mean accuracy over the four samples) and as pass@4 (a problem counts as solved if any of its four samples is correct).

\section{Results}\label{sec:results}

\begin{figure}[t]
\centering
\begin{minipage}[b]{0.52\linewidth}
\centering
\footnotesize
\setlength{\tabcolsep}{4pt}
\begin{tabular}{@{}lcccc@{}}
\toprule
 & \multicolumn{2}{c}{Gate} & \multicolumn{2}{c}{Test} \\
\cmidrule(lr){2-3}\cmidrule(l){4-5}
Model & avg@4 & $\Delta_{\gate}$ & avg@4 & pass@4 \\
\midrule
$\Mz$ (Qwen3.8-27B) & 73.95 & -- & 76.30 & 82.60 \\
$M_1$ & 75.15 & +1.20 & -- & -- \\
$M_2$ & 75.65 & +0.50 & -- & -- \\
$M_3$ & 75.75 & +0.10 & -- & -- \\
$M_4$ & 76.45 & +0.70 & -- & -- \\
$M_5$ (final) & \textbf{76.45} & 0.00 & \textbf{77.75} & \textbf{84.80} \\
\midrule
$M_5 - \Mz$ & \textbf{+2.50} & & \textbf{+1.45} & \textbf{+2.20} \\
\bottomrule
\end{tabular}
\captionof{table}{\textbf{Self-editing raises gate accuracy by 2.50 points, and the final model gains 1.45 points avg@4 and 2.20 points pass@4 on the held-out test split.} Gate accuracy after each generation and final test accuracy (\%). $\Delta_{\gate}$: paired gain over the previous model on the gate set (points). The test split is reported for $\Mz$ and the final model.}
\label{tab:main}
\end{minipage}\hfill
\begin{minipage}[b]{0.45\linewidth}
\centering
\includegraphics[width=\linewidth]{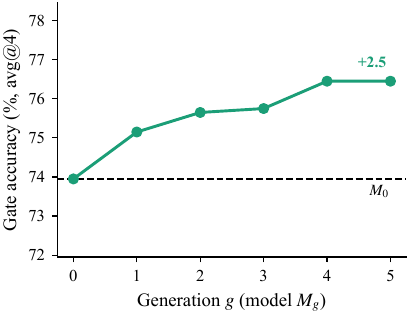}
\caption{\textbf{Gate accuracy rises from 73.95\% ($\Mz$) to 76.45\% ($M_5$).} Gate accuracy (\%, avg@4) of the current model $\Mg$ after each generation; dashed: $\Mz$.}
\label{fig:gens}
\end{minipage}
\end{figure}

\subsection{Gains across generations}\label{sec:results-gens}
\Cref{tab:main} and \cref{fig:gens} follow the current model through five generations of \ours{}.
Each generation the edited model writes the next edit, and an edit is merged only when it beats the model that wrote it on the hidden gate, so every accepted step is a paired gain the model produced by editing itself.
\ours{} raises gate accuracy from 73.95\% for $\Mz$ to 76.45\% for the final model $M_5$, a cumulative gain of 2.50 points.
The accepted edits gain +1.20, +0.50, +0.10 and +0.70 points. The first edit brings the largest step, and every later generation, written by a model that has already edited itself, still finds a further improvement; the fourth generation, written by $M_3$, brings the second-largest step: the edited model remains an effective editor of its own weights.

\subsection{Held-out test accuracy}\label{sec:results-test}
The test split is never screened, gated or shown to the editor, so it measures whether the self-written edits generalise beyond the problems that selected them.
On these 500 problems, avg@4 accuracy rises from 76.30\% for $\Mz$ to 77.75\% for $M_5$ (+1.45 points) and pass@4 from 82.60\% to 84.80\% (+2.20 points; \cref{tab:main}) without a single gradient step.
The edited model both answers more reliably per sample and solves problems that $\Mz$ missed in all four samples, so the self-written edits carry over to problems that played no part in selecting them.

\section{Conclusion}\label{sec:conclusion}
We introduced gradient-free recursive self-editing, in which a language model writes the program that changes its own weights and the edited model writes the next change.
\ours{} grounds each edit in the model's own reasoning on fresh problems, explores its weights in an agentic session with a white-box toolkit, accepts an edit only on a hidden paired gate, and merges it so the improved model becomes the next editor.
On Qwen3.8-27B, five generations raise gate accuracy by 2.50 points (73.95\% to 76.45\%) and held-out test accuracy by 1.45 points avg@4 (76.30\% to 77.75\%) and 2.20 points pass@4 (82.60\% to 84.80\%) without a single gradient step.

\paragraph{Discussion.}
The loop needs no reward model, no gradient and no human-written edit: the model decides where and how to change its weights from its own answers, and a fixed paired gate decides whether the change is kept.
Every gain in \cref{tab:main} is therefore a change the model chose for itself, stored as a short program and a low-rank tensor that can be read, rerun or reverted.
Because the improved model writes the next edit, the editor improves together with the model it edits.
The cost of a generation is dominated by sampling the screen and grading eight candidates, so the loop scales with inference rather than with training.

\paragraph{Future work.}
Next steps are self-editing across several domains at once, longer horizons with more generations, larger editors, and richer edit spaces beyond low-rank updates, such as edits that change several modules jointly.

\clearpage
\bibliographystyle{plainnat}
\bibliography{refs}

\appendix
\section{Method Details of \ours{}}\label{sec:app-method}

\subsection{Tools}\label{sec:app-tools}
\Cref{tab:app-tools} lists the program tools with their costs; every tool takes keyword arguments and returns handles to objects held by the trusted runtime.
Exploration through \texttt{run\_python} uses the same tools, and GPU tools draw on a separate 600-second exploration budget per generation, shared by the eight sessions; no tool evaluates a candidate.

\begin{table}[h]
\centering
\small
\caption{Program tools. ``Free'' tools cost nothing; GPU tools draw on the generation's 7,200-second GPU budget, shared by its eight programs; \texttt{tensor\_compute} runs on CPU within the program's wall clock.}
\label{tab:app-tools}
\begin{tabular}{@{}p{0.36\linewidth}p{0.48\linewidth}l@{}}
\toprule
Tool & Purpose & Cost \\
\midrule
\texttt{screen\_results(ids, filter, limit, offset)} & own screen answers: statement, gold, extracted answers, scores & free \\
\texttt{screen\_output(id, sample, offset)} & one full answer, reasoning included, paged & free \\
\texttt{train\_ids}, \texttt{train\_example(id)} & train problems with gold answers & free \\
\texttt{modules}, \texttt{tensors}, \texttt{output\_matrices} & module and tensor names and shapes & free \\
\texttt{diagnostics(module, layer, top, sort)} & per-module norms, singular values, merged-edit norm, effective rank, intruder score, relative change & free \\
\texttt{svd(module, k)} & top-$k$ singular triplets of the current weight & free \\
\texttt{acts(module, ids, k, source)} & activation means and principal directions on train prompts, screen prompts or own answers & GPU \\
\texttt{new\_lora}, \texttt{rank1}, \texttt{from\_direction}, \texttt{scale\_singular} & closed-form low-rank constructions & free \\
\texttt{unit}, \texttt{direction}, \texttt{project\_out} & vector algebra; $W \leftarrow W - \alpha uu^\top W$ as a rank-one term & free \\
\texttt{tensor\_compute(code, tensors, vectors)} & sandboxed CPU PyTorch on up to four FP32 tensors & CPU time \\
\texttt{scale}, \texttt{interpolate}, \texttt{combine}, \texttt{drop}, \texttt{truncate\_rank}, \texttt{intruder\_scale}, \texttt{relchange\_prune}, \texttt{orthogonalize}, \texttt{ties} & LoRA operators & free \\
\texttt{stack}, \texttt{candidate(cid)} & merged edits of earlier generations; earlier candidates & free \\
\texttt{save\_edit}, \texttt{load\_edit}, \texttt{list\_edits}, \texttt{drop\_edit} & keep edits between exploration and the final program & free \\
\texttt{inspect(lora)} & modules, rank, cap check & free \\
\texttt{train\_loss(lora, ids, n)} & forward-only answer NLL on train problems vs.\ without the edit & GPU \\
\texttt{compare(cid, other)} & fixed / broken problems between evaluated models (ids and statistics) & free \\
\bottomrule
\end{tabular}
\end{table}

\subsection{What the editor sees}\label{sec:app-context}
The system prompt describes the setting (the editor is the model being edited; edits are merged if they win on the hidden gate), the caps, the toolkit with signatures, the number of tool rounds, and that the session is one of eight independent sessions.
Every message carries the summary of the current model's screen answers, the diagnostics table, the gate outcomes of earlier generations' candidates (accuracy, paired gain and interval, fixed and broken counts; \texttt{search\_history} and \texttt{read\_record} also list the ids of fixed and broken problems, never their content), and the best programs of earlier generations.
Long sessions are compacted: older tool outputs and reasoning are elided once the prompt exceeds 60\% of the 131,072-token context.

\subsection{Sandbox and data boundaries}\label{sec:app-sandbox-b}
Programs and \texttt{run\_python} code run in a separate process with memory and time limits and no network access; tools are reached only through the \texttt{api} object, and objects are referred to by handles.
Activation tools accept only train or screen ids, the statements and answers of gate and test problems are never readable, and the final program must be self-contained: it is re-run from scratch and its output is checked before grading.

\section{Hyper-parameters and Compute}\label{sec:app-hparams}
\Cref{tab:hparams} lists the settings of the main run (math).

\begin{table}[h]
\centering
\footnotesize
\renewcommand{\arraystretch}{0.92}
\caption{Hyper-parameters of \ours{} (main run).}
\label{tab:hparams}
\begin{tabular}{@{}p{0.34\linewidth}p{0.60\linewidth}@{}}
\toprule
Setting & Value \\
\midrule
\multicolumn{2}{@{}l}{\emph{Model and decoding}} \\
Model $\Mz$ & Qwen3.8-27B, BF16 \\
Editor & current model $\Mg$, thinking mode; temperature 0.6, top-$p$ 0.95, top-$k$ 20 \\
Editor output caps & 16,384 tokens per tool turn and for the final turn; context 131,072 tokens \\
Editor session limits & 30 tool rounds, 100 tool calls, 160,000 completion tokens, 2,400\,s per candidate \\
Screen / gate / test decoding & thinking mode, temperature 0.6, top-$p$ 0.95, top-$k$ 20, $\le$16,384 tokens, 4 samples (avg@4), seed 42 \\
Grading & deterministic math answer equivalence \\
\midrule
\multicolumn{2}{@{}l}{\emph{Data}} \\
Pool & DAPO-Math-17K, 14,738 problems after decontamination and de-duplication \\
Gate set $\gate$ & 500 problems, fixed, never readable \\
Test set & 500 problems, fixed, reported only \\
Screen set & 500 problems, drawn afresh each generation \\
Train set & remaining 13,238 problems per generation (statements and gold answers readable) \\
\midrule
\multicolumn{2}{@{}l}{\emph{Edits}} \\
LoRA rank cap & 64 per module \\
Norm cap $\tau$ & $\|\Delta W\|_F/\|W_g\|_F \le 0.1$ per module \\
Merge & $W' = \mathrm{bf16}(W + BA)$, FP32 accumulation, edited shards only \\
\midrule
\multicolumn{2}{@{}l}{\emph{Search and gate}} \\
Sessions per generation & 8 independent sessions, 1 candidate each \\
Acceptance & largest strictly positive paired gain on $\gate$ \\
Generations & 5; stop after 3 consecutive generations without acceptance \\
GPU budget & per generation, shared by the 8 sessions: 7,200\,s for programs, 600\,s for exploration \\
\texttt{tensor\_compute} & $\le$4 tensors per call, 300\,s, 16\,GB, CPU only \\
Program sandbox & 1,800\,s, 16\,GB, no network \\
\midrule
\multicolumn{2}{@{}l}{\emph{Hardware}} \\
GPUs & 8 per run: 7 vLLM replicas of $\Mg$ (editor, scoring, LoRA hot-loading) + 1 tool GPU (SVD, activations, losses, merge) \\
\bottomrule
\end{tabular}
\end{table}

\end{document}